%% file: acl_latex.tex
\pdfoutput=1

\documentclass[11pt]{article}

\usepackage[preprint]{acl}

\usepackage{times}
\usepackage{latexsym}

\usepackage[T1]{fontenc}

\usepackage[utf8]{inputenc}

\usepackage{microtype}

\usepackage{inconsolata}

\usepackage{graphicx}

\usepackage{multirow}
\usepackage{amsmath}
\usepackage{comment}
\usepackage{hyperref}
\usepackage[colorinlistoftodos]{todonotes}

\input{macro}

\title{CORD: Balancing  COnsistency and Rank Distillation\\
for Robust Retrieval-Augmented Generation}

\author{\textbf{
Youngwon Lee\textsuperscript{*}\quad
Seung-won Hwang\thanks{Work done while visiting Snowflake. Correspondence to: \href{mailto:seungwonh@snu.ac.kr}{\texttt{seungwonh@snu.ac.kr}}.}\quad
Daniel Campos}\\
\textbf{
Filip Graliński\quad
Zhewei Yao\quad
Yuxiong He
}\\
Snowflake AI Research\qquad Seoul National University\textsuperscript{\rm *}\\
}

\begin{document}
\maketitle

\input{sec/0_abs}

\input{sec/1_intro_new2}

\input{sec/3_method}

\input{sec/4_results}

\input{sec/5_conclusion}

\bibliography{anthology_part1,anthology_part2,custom}

\clearpage
\appendix
\input{sec/2_relwork}

\input{sec/6_appendices}

\end{document}

%% file: macro.tex
\usepackage{comment}
\usepackage{amsmath}
\usepackage{pifont}
\usepackage{caption, booktabs}
\usepackage{algorithm}
\usepackage{algpseudocode}
\usepackage{amssymb}
\usepackage{setspace}
\usepackage{bbm}
\usepackage{arydshln}
\usepackage{afterpage}
\usepackage{enumitem}

\makeatletter
\newcommand\footnoteref[1]{\protected@xdef\@thefnmark{\ref{#1}}\@footnotemark}
\makeatother
\newcolumntype{P}[1]{>{\centering\arraybackslash}p{#1}}

\usepackage{xspace}
\newcommand{\ours}[0]{\textsc{CORD}\xspace}

\usepackage{graphicx}
\usepackage{adjustbox}
\usepackage{subfig}
\makeatletter
\newcommand{\thickhline}{
    \noalign {\ifnum 0=`}\fi \hrule height 1pt
    \futurelet \reserved@a \@xhline
}
\newcolumntype{"}{@{\hskip\tabcolsep\vrule width 1pt\hskip\tabcolsep}}
\makeatother
\usepackage{mathabx}
\usepackage{amsfonts}

\makeatletter
\newcommand*{\blackleq}{
  \mathrel{
    \mathpalette\@blackleq{}
  }
}
\newcommand*{\@blackleq}[2]{
  \vcenter{
    \m@th
    \setbox0=\hbox{$#1\mkern3mu$}
    \setbox2=\hbox{$#1\vcenter{}$}
    \setbox4=\hbox{\raisebox{-\ht2}[.2pt][.2pt]{$#1-$}}
    \hbox{$#1\blacktriangleleft$}
    \nointerlineskip
    \kern\wd0 
    \copy4 
  }
}
\makeatother
\usepackage{dsfont}
\newcommand{\argmax}{\operatornamewithlimits{argmax\text{ }}}

\usepackage{multirow}

\usepackage{mathtools}

\usepackage{tablefootnote}

\usepackage{placeins} %

\usepackage{tikz}

%% file: sec/0_abs.tex
\begin{abstract}

With the adoption of retrieval-augmented generation (RAG), large language models (LLMs) are expected to ground their generation to the retrieved contexts.
Yet, this is hindered by position bias of LLMs, failing to evenly attend to all contexts.
Previous work has addressed this by synthesizing contexts with perturbed positions of
gold segment, creating a position-diversified train set.
We extend this intuition to propose consistency regularization with augmentation and distillation.
First, we augment each training instance with its position perturbation to encourage consistent predictions, regardless of ordering.
We also distill behaviors of this pair, although it can be
counterproductive in certain RAG scenarios where the given order from the retriever is crucial for generation quality.
We thus propose CORD, balancing COnsistency and Rank Distillation.
CORD adaptively samples noise-controlled perturbations from an interpolation space, ensuring both consistency and respect for the rank prior.
Empirical results show this balance enables CORD to outperform consistently in diverse RAG benchmarks.

\end{abstract}

%% file: sec/1_intro_new2.tex
\section{Introduction}

\label{sec:intro}

Recently, large language models (LLMs) have incorporated retrievers to augment input contexts for more grounded generation.
However, during retrieval-augmented generation (RAG), LLMs reportedly suffer from position bias
where
they pay disproportionate attention to different parts,
worsened as the input becomes longer
\citep{liu-etal-2024-lost}.
An existing solution has synthesized a training set by randomizing the position of gold segment~\citep{an-etal-2024-make}.
It allows LLMs to implicitly learn that relevant information can appear at any position, mitigating position bias.

\begin{figure}[t]
  \centering
  \includegraphics[width=.95\linewidth]{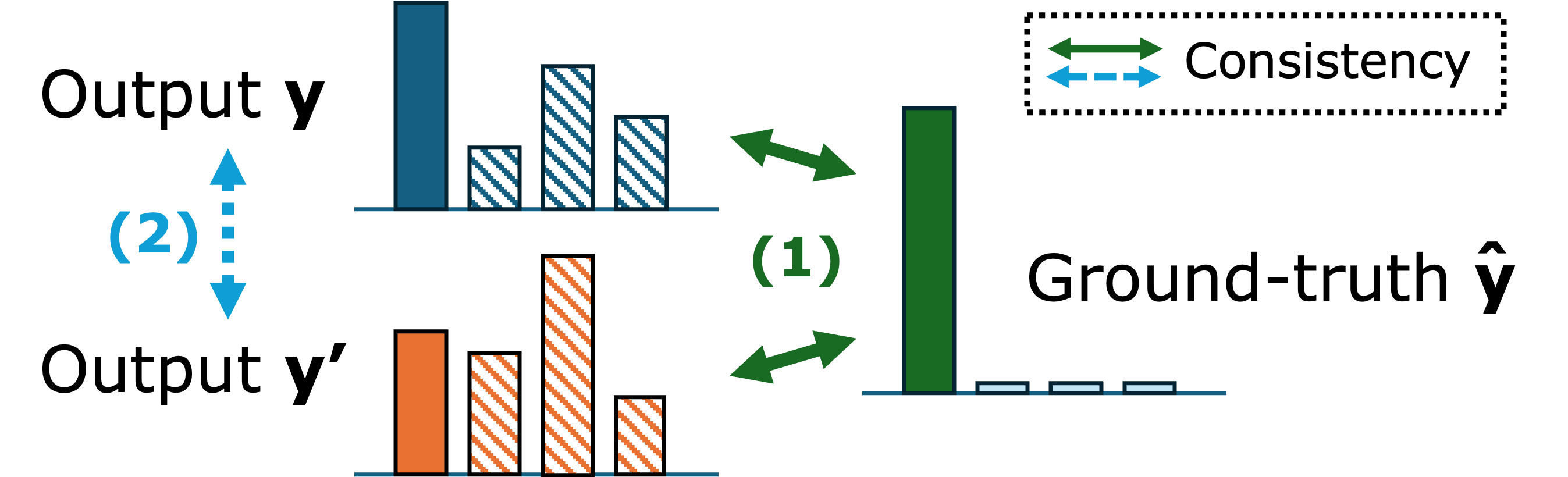}
  \caption{
  Enforcing consistency with (1) augmentation
  (green) and (2) distillation (blue).
  }
  \label{fig:concept_agg}
\end{figure}

Our distinction is to pursue dual goals of (1) \textbf{CO}nsistency for mitigating position bias and (2) \textbf{R}ank \textbf{D}istillation, learning to utilize meaningful signals in the given order from the retriever and also to denoise it, for robust RAG.

For CO, we extend the position-perturbing training intuition, by
augmenting the retriever-given order $\mathbf{c}$ with its perturbation $\mathbf{c}^\prime$,
sharing the same ground truth $\hat{y}$.
Green arrows in Figure~\ref{fig:concept_agg} visualize how this augmentation
 indirectly enforces consistency
 by guiding predictions
$y$ from $\mathbf{c}$ and $y^\prime$ from $\mathbf{c}^\prime$, to converge to the ground-truth $\hat{y}$.

Another more direct approach is adding
a distillation loss penalizing the distributional divergence in all outputs.
The blue arrow in Figure~\ref{fig:concept_agg}
visualizes
this loss further
incentivizing consistent internal representation, by distilling
 `dark knowledge'~\citep{dark1, dark2, dark3} from one to another.

\input{tab/tab_risk}

However, pursuing CO objective alone, without balancing it with the RD objective,
is counterproductive in some scenarios as illustrated in Table~\ref{tab:risk}.
It contrasts two representative real-life RAG scenarios A and B:~\footnote{For presentation brevity, we reveal in later section.}
In A, retriever provides a reliable rank prior, such that distilling predictions from a randomized ordering can unlearn this helpful prior, as
evidenced by a decrease in generation quality after consistency regularization.
Meanwhile, in B, where generation is not sensitive to the given order, CO objective enhances performance.

Our technical contribution is to adapt
$\mathbf{c}^\prime$ to the given scenario, by controlling the degree of perturbation, in place of
$\mathbf{c}^\prime$ with a fixed randomization.
We define an interpolated space of perturbations and dynamically sample
an appropriate level of perturbation from it.
Table~\ref{tab:risk} shows \ours
outperforms in both scenarios, by sampling
smaller perturbations in scenario A, where rank prior is important, and larger perturbations in scenario B, where robustness to position bias is crucial.

Our contribution can be summarized as follows: (1) We propose \ours, balancing connsistency and rank distillation in RAG. (2) We show distilling with a controlled perturbation, sampled from an interpolated space of teachers, is effective across 5 diverse RAG scenarios, whereas existing consistency methods fall short.

%% file: tab/tab_risk.tex
\begin{table}[t]
    \centering
    \fontsize{10.5pt}{12pt}\selectfont
    \begin{tabular}{lcc}
        \thickhline
        Method 
        & (A) & (B)
        \\ \hline
        Given order & 41.34 \phantom{($\uparrow$)} & 56.52 \phantom{($\downarrow$)} \\
        \phantom{0} + consistency & 36.87 (\textcolor{red}{$\downarrow$}) & 57.87 (\textcolor{cyan}{$\uparrow$}) \\
        \rowcolor{gray!10} \ours (ours) & \textbf{44.74} (\textcolor{cyan}{$\uparrow$}) & \textbf{58.71} (\textcolor{cyan}{$\uparrow$}) \\
        \thickhline
    \end{tabular}
    \caption{
    Representative
    RAG scenarios A and B, where distillation may hinder or enhance, respectively.
    }
    \label{tab:risk}
\end{table}

%% file: sec/3_method.tex
\input{tab/tab_main}

\begin{figure}[t]
  \centering
  \includegraphics[width=\linewidth]{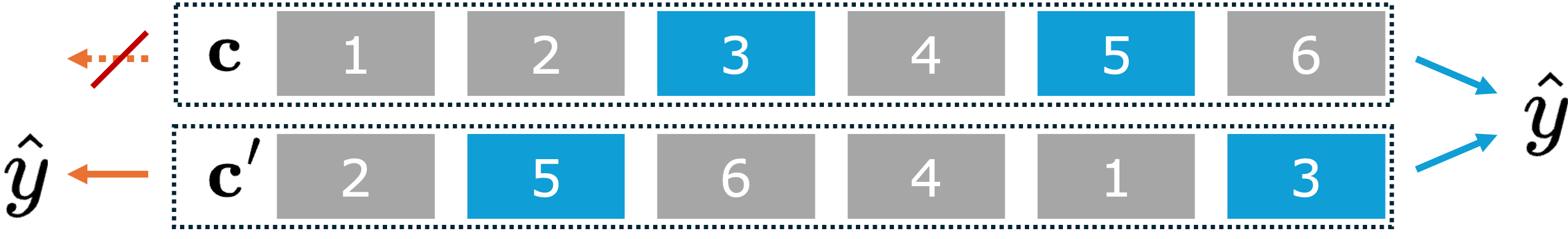}
  \caption{
   (Left) \textsc{In2} only uses $\mathbf{c}^\prime$.
   (Right) 
  We augment the given order $\mathbf{c}$ (top)
  with perturbed ranking $\mathbf{c}^\prime$ (bottom) and use both.
  }
  \label{fig:data_aug}
\end{figure}

\section{Method}

\subsection{CO: Consistency regularization}

\label{subsec:consistency}

We propose to mitigate position bias by regularizing output consistency over possible perturbations,
through (1) augmentation and (2) distill loss.

First, we explain how augmenting position-perturbed examples
contributes to consistency.
We first formalize
RAG as generating an answer $y$ given an input $x$,
\begin{equation} \label{eq:task_description}
y \sim p( \cdot \,|\, x, \mathbf{c}),
\end{equation}
along with the sequence of $n$ retrieved contexts $\mathbf{c}=[c_1; c_2; \cdots; c_n]$.
Then, for a training triplet $(x, \mathbf{c}, \hat{y})$ 
the negative log-likelihood (NLL) loss for maximum likelihood estimation training is
\begin{equation} \label{eq:nll_loss}
\mathcal{L}_{\textrm{n}} = -\sum_{t} \log p( \hat{y}_t \,|\, x, \mathbf{c} , \hat{y}_{<t}),
\end{equation}
which encourages the model to produce the correct answer $\hat{y}$ given the input $x$ and retrieved contexts.

Inspired by \citet{an-etal-2024-make}, referred to as
\textsc{In2},
we employ position perturbation to
augment $\mathbf{c}$ from the corpus $\mathcal{C}$ with
$\mathbf{c}^\prime$.
For comparison, \textsc{In2} synthesized question and context $(q,\mathbf{c})$ pairs where the gold passage $s$ for generating the gold answer $\hat{y}$ appears in various positions.
As Figure~\ref{fig:data_aug} shows,
we retain both the original $(q,\mathbf{c},\hat{y})$ and the perturbed examples $(q,\mathbf{c}^\prime,\hat{y})$:
Unlike \textsc{In2}'s using $\mathbf{c}^\prime$ only for training (orange arrows),
we train over the augmented dataset $\mathcal{C}^\prime$ %
which includes both $\mathbf{c}$ and $\mathbf{c}^\prime$ (blue arrows). Predictions for both are supervised to converge to the same ground-truth $\hat{y}$ using the loss in Eq~\ref{eq:nll_loss}.

Second, by adding a distill loss, 
we can further match token-level output probability distributions for $\mathbf{c}$ and $\mathbf{c}^\prime$.
We use the sum of Jensen-Shannon Divergence (JSD) between output probability distributions at each time step $t$ for this purpose:\footnote{While we default to summing all terms,
the number of time steps $t$ to aggregate in Eq~\ref{eq:cons_loss} can be adjusted for efficiency, as detailed in Appendix~\ref{app:first_only}.} 
\begin{equation} \label{eq:cons_loss}
\mathcal{L}_{\textrm{d}} = \sum_{t} \textrm{JSD} \left( f_t(\mathbf{c}) \,\Vert\, f_t(\mathbf{c}^\prime) \right),
\end{equation}
where $f_t(\mathbf{c}) = p( \hat{y}_t \,|\, x, \mathbf{c}, \hat{y}_{<t})$.
This
encourages the model to align its internal representations of input and association with the output, encoded in the `dark knowledge'~\citep{dark1, dark2, dark3} across different perturbations.

Finally, the two types of loss in Eq~\ref{eq:nll_loss} and \ref{eq:cons_loss} can be combined to obtain our training objective:
\begin{equation} \label{eq:total_loss}
\mathcal{L} = \mathcal{L}_{\textrm{n}} + \lambda\cdot\mathcal{L}_{\textrm{d}},
\end{equation}
where the hyperparameter $\lambda$ determines the relative strength of the two terms.

\subsection{RD: Adaptive teacher selection for rank distillation}
\label{subsec:ours}

However, as previously outlined in Table~\ref{tab:risk}A,
distill loss on a random perturbation $\mathbf{c}^\prime$ may interfere with the RD objective
in an RAG scenario where retriever provides a meaningful ranking $\mathbf{c}$ with valuable prior: In this work, we consider
MS MARCO~\citep{bajaj-etal-2018-ms} as a representative example, where an industry-scale complex retrieval system provides the ranking.

\begin{figure}[t]
  \centering
  \includegraphics[width=\linewidth]{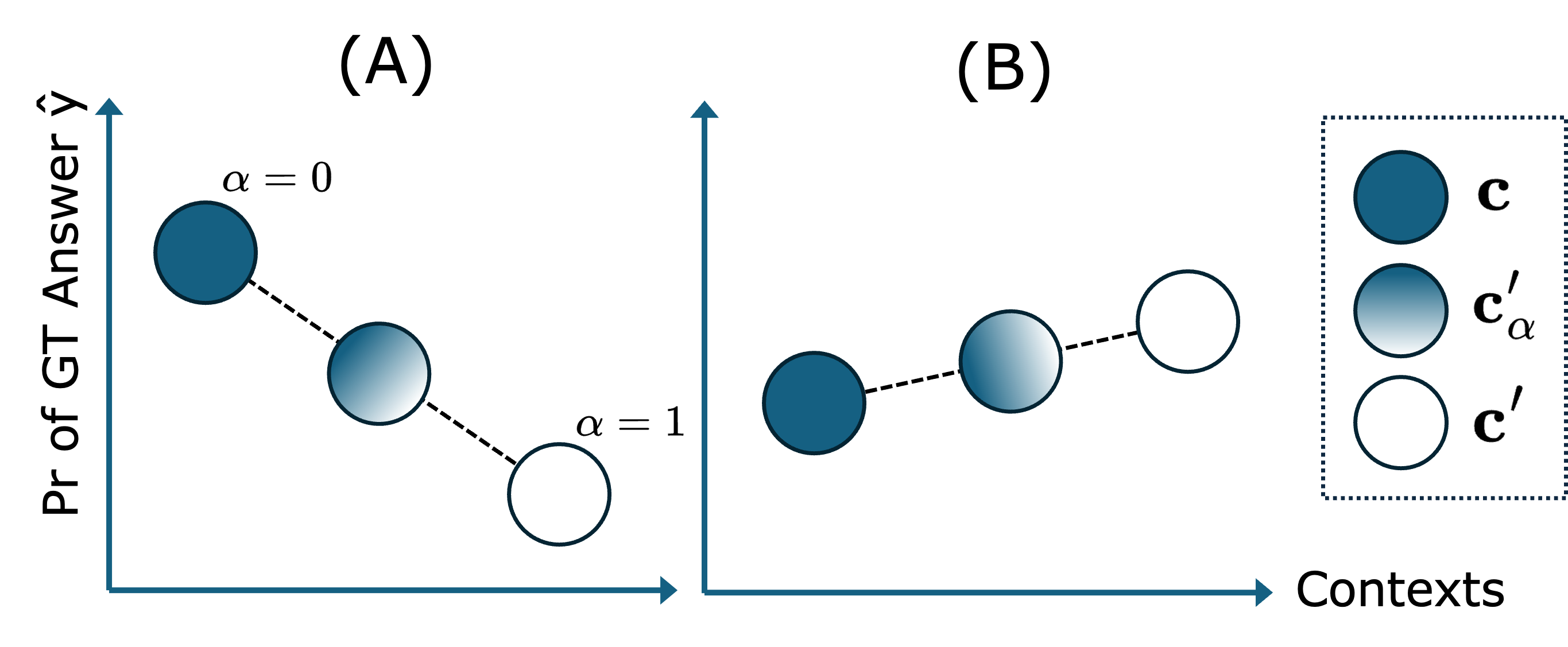}
  \caption{
  Interpolated sample space for scenario A and B from Table~\ref{tab:risk}, where (A, left) perturbation leads to a large drop in probability of ground-truth $\hat{y}$, and (B, right) with no such drop.
  }
  \label{fig:choice}
\end{figure}

Figure~\ref{fig:choice}A depicts such unlearning of ranker prior, when distilled from a random perturbation in scenario A.
The $y$-axis in the plot represents the probability the LLM assigns to the ground-truth answer, $p(\hat{y}\,|\,x,\mathbf{c})$ for the given order $\mathbf{c}$ (solid  circle) and $p(\hat{y}\,|\,x,\mathbf{c}^\prime)$ for random perturbation $\mathbf{c}^\prime$ (empty circle).
In MS MARCO, the given order $\mathbf{c}$ carries a useful prior, resulting in high probability of the ground-truth $p(\hat{y}\,|\,x,\mathbf{c})$.
Randomizing this order would greatly lower the probability $p(\hat{y}\,|\,x,\mathbf{c}^\prime)$, such that enforcing consistency between the two would unlearn the benefit of rank prior.

To tackle this,
instead of fixing $\mathbf{c}^\prime$ as a random perturbation,
we define a sample space and strategy for adaptive teacher selection,
to control the degree of perturbation for distillation.
We introduce an interpolation of $\mathbf{c}$ and $\mathbf{c}^\prime$ with a controlled noise degree of  $\alpha$, denoted as 
$\mathbf{c}^\prime_\alpha$:
Here, the lower ranked $\alpha$ proportion of the retrieved contexts is randomized while the remaining retains the given order.
In Figure~\ref{fig:choice}, such interpolated sample is shown as
a shaded circle on a dotted line,
 the interpolated path connecting $\mathbf{c}$ and $\mathbf{c}^\prime$, 
 as the noise degree $\alpha$ varies from 0 to 1.
For brevity, we assume
a desirable single value of $\alpha$ for the given task
is known a priori, and later discuss how to find it in Section~\ref{subsec:adaptive_alpha}.

This interpolation allows to select a better teacher between $\mathbf{c}^\prime_\alpha$ and $\mathbf{c}^\prime$ by choosing the one with a higher probability of predicting the ground truth.
As shown in Figure~\ref{fig:choice}A, small perturbations tend to yield higher $y$ values in scenario A as they retain the given order in part,
leading to $\mathbf{c}^\prime_\alpha$ chosen for distillation.
This corresponds to
ensembling two retrievers,
which agree on top-ranked documents but diversify the ranks of the rest.

An added advantage is, the same approach
seamlessly supports scenario B, where there is no conflict between CO and RD.
As illustrated in Figure~\ref{fig:choice}B, the $y$-axis score remains relatively stable across different orderings, and moreover, the score is no longer sensitive to ordering.
Thus, pairing the given order with the one that has a higher $y$ score essentially serves the goal of pursuing CO.

\subsection{Score-aware teacher sampling}

\label{subsec:adaptive_alpha}

So far, we have mainly focused on utilizing \emph{rank} prior from the retriever; however, the retriever may provide varying level of information in different RAG scenarios, such as score for each item as well.
We describe how to incorporate such additional signals into adaptive teacher sampling.

When no prior knowledge of the distribution of the probability of ground-truth $p(\hat{y}\,|\,x,\mathbf{c}^\prime_\alpha)$ over the interpolated path is known,
we follow the principle of maximum entropy~\citep{jaynes-1957-information}
to assume uniform distribution.
That is, we choose to sample $\alpha=0.5$ from the interpolated space defined in Section~\ref{subsec:ours}, where $\alpha$ varies in the range of $(0,1)$.

Alternatively, we utilize retriever scores as a proxy for the unknown distribution of $p(\hat{y}\,|\,x,\mathbf{c}^\prime_\alpha)$, from which the optimal noise level $\alpha$ can be determined.
Specifically, we aim to extract the most confident top-ranked contexts identified by the retriever, by preserving the contexts ranked above the largest discontinuity in scores and perturbing the rest.
Given scores $s_i$ for each retrieved context $c_i\in \mathbf{c}$,
which are sorted in descending order of score, i.e., $s_1 > s_2 > \cdots > s_n$, we locate the adjacent pair of passages with the largest gap in retriever score
$\hat{i} = \argmax_i (s_i - s_{i+1})$
and perturb the passages ranked lower than $\hat{i}$.
In other words, we choose $\alpha=1-\hat{i}/n$ for this example.
Intuitively, this approximates finding the largest acceptable degree of noise that would still result in sufficiently high $p(\hat{y}\,|\,x,\mathbf{c}^\prime_\alpha)$.

%% file: tab/tab_main.tex
\afterpage{
\begin{table*}[htp]
    \centering
    \fontsize{10.5pt}{12pt}\selectfont
    \begin{tabular}{lcccccccc}
        \thickhline
        & \multicolumn{2}{c}{\multirow{1.25}{*}{MS MARCO}}
        & \multicolumn{2}{c}{\multirow{1.25}{*}{HotpotQA}}
        & \multicolumn{2}{c}{\multirow{1.25}{*}{NQ}}
        & \multicolumn{1}{c}{\multirow{1.25}{*}{MN}}
        & \multicolumn{1}{c}{\multirow{1.25}{*}{MN-IDK}}
        \\
        \cmidrule(r){2-3}
        \cmidrule(r){4-5}
        \cmidrule(r){6-7}
        \cmidrule(r){8-8}
        \cmidrule(r){9-9}
        \multirow{-1.25}{*}{Finetuning Objective} 
        & \multicolumn{1}{c}{\multirow{-1.25}{*}{R-L}}
        & \multicolumn{1}{c}{\multirow{-1.25}{*}{GPT-4}}
        & \multicolumn{1}{c}{\multirow{-1.25}{*}{EM}}
        & \multicolumn{1}{c}{\multirow{-1.25}{*}{GPT-4}}
        & \multicolumn{1}{c}{\multirow{-1.25}{*}{Acc}}
        & \multicolumn{1}{c}{\multirow{-1.25}{*}{GPT-4}}
        & \multicolumn{1}{c}{\multirow{-1.25}{*}{F1}}
        & \multicolumn{1}{c}{\multirow{-1.25}{*}{Acc}}
        \\\hline
        No finetuning 
            & {41.34} & {51.94} 
            & {42.86} & {66.50}
            & {52.18} & {62.46} 
            & {56.52} & {54.82} 
            \\
        $\mathcal{L}_{\textrm{nll}}$ on $\mathcal{C}^\prime$ 
            & {44.52} & {\textbf{57.28}}
            & {58.62} & {83.75}
            & {55.60} & {63.51}
            & {56.25} & {95.78}
            \\
        \rowcolor{gray!10}
        \ours
            & {\textbf{44.74}} & {\textbf{57.28}}
            & {\textbf{63.55}} & {\textbf{85.72}}
            & {\textbf{58.55}} & {\textbf{63.72}} 
            & {\textbf{58.71}} & {\textbf{98.83}}
            \\
        \thickhline
    \end{tabular}
    \caption{
    RAG performance with Phi-3 3B as the generator and different finetuning strategies applied.
    }
    \label{tab:main}
\end{table*}
}

%% file: sec/4_results.tex
\input{tab/tab_mleonly}
\input{tab/tab_scoreaware}

\section{Results}

We design evaluations  to answer these research questions:
(RQ1) Does \ours pursue dual goals of CO and RD effectively?
(RQ2) Does \ours adaptively choose $(\mathbf{c}, \mathbf{c}^\prime)$ pair in different scenarios?
(RQ3) How can the noise degree $\alpha$ for interpolation be tuned per task or example?

\subsection{Experimental settings}

We have evaluated our proposed method on several QA benchmarks: MS MARCO~\citep{bajaj-etal-2018-ms}, HotpotQA~\citep{yang-etal-2018-hotpotqa}, NaturalQuestions (\citet{kwiatkowski-etal-2019-natural}; NQ) as reorganized by \citet{liu-etal-2024-lost}.
We further consider multi-needle (MN) dataset, which is built following \citet{an-etal-2024-make},
as a scenario where irrelevant contexts are prevalent and retriever prior is not meaningful.\footnote{This corresponds to scenario B in Table~\ref{tab:risk} and Figure~\ref{fig:choice}.}

For evaluation, we used widely reported metrics for each benchmark, namely ROUGE-L for MS MARCO, exact match (EM) for HotpotQA, and span-based exact match, or `accuracy' for NQ.
We also adopted the evaluation protocol from \citet{yang-etal-2024-crag} using GPT-4, allowing more flexibility in answers.
For MN where answers typically contain a few sentences, we report sentence-level F1,
and for MN-IDK,
an unanswerable split of MN,
we report accuracy.
Further details can be found in Appendix~\ref{app:impl_details}.

\begin{figure}[t]
  \centering
  \includegraphics[width=.85\linewidth]{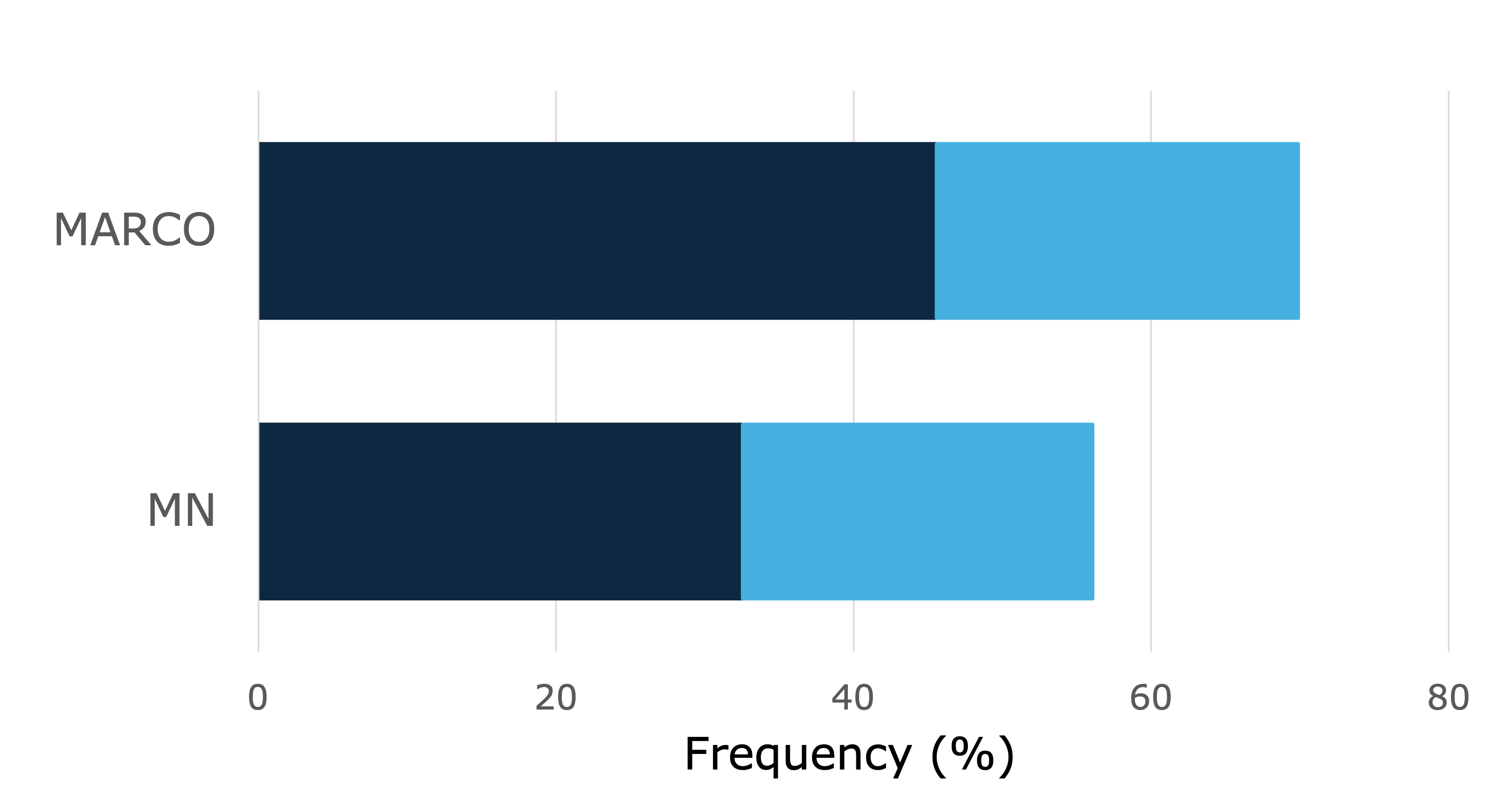}
  \caption{
  (Top) On MS MARCO, the interpolated noise-controlled perturbation $\mathbf{c}^\prime_\alpha$ (dark blue) is much more likely to be paired with the given order $\mathbf{c}$, than $\mathbf{c}^\prime$ (light blue). (Bottom) The gap is much smaller on MN.
  }
  \label{fig:ratio}
\end{figure}

\subsection{Results}

\paragraph{Bias mitigation and rank distillation} Table~\ref{tab:main} shows that our proposed method outperforms the baselines
across all benchmarks, validating its effectiveness in pursuing dual goals of CO and RD.

In addition, Table~\ref{tab:mleonly} shows the importance of denoising through consistency in rank distillation.
There is a clear performance gap between the model trained on the given order $\mathbf{c}$ without augmentation (2nd row), and those augmented (3rd \& 4th) on MS MARCO.
This suggests that even with a strong rank prior, consistency across slight perturbation positively contributes to RD, by mitigating potential bias from retriever or generator.

\paragraph{Adaptive pair selection}
\ours indeed selects the proper teacher for enforcing consistency,
while the tendency in choices exhibit clear difference per different RAG scenario,
as shown in Figure~\ref{fig:ratio}.
The ratio of $\mathbf{c}^\prime_\alpha$ paired with $\mathbf{c}$ is shown with dark blue, while the ratio of $\mathbf{c}^\prime$ paired with $\mathbf{c}$ is presented by light blue bar.
Comparing MS MARCO (top) and MN (bottom), it is clearly shown that $\mathbf{c}^\prime_\alpha$ is much more likely to be paired with $\mathbf{c}$ in the former,
where the RD objective is more prominent.
This supports our rationale behind designing adaptive teacher selection in Section~\ref{subsec:ours}.

\paragraph{Score-aware teacher sampling}
Table~\ref{tab:scoreaware} shows that score-aware dynamic adjustment of $\alpha$, described in Section~\ref{subsec:adaptive_alpha} brings further gain; the effective mean value of $\alpha$ throughout the train set was 0.8, suggesting a larger portion of the ranking was allowed to be perturbed.

%% file: tab/tab_mleonly.tex
\afterpage{
\begin{table}[t]
    \centering
    \fontsize{10.5pt}{12pt}\selectfont
    \begin{tabular}{lcc}
        \thickhline
        & \multicolumn{2}{c}{\multirow{1.25}{*}{MS MARCO}} \\
        \cmidrule(r){2-3}
        Finetuning Method 
        & \multirow{-1.25}{*}{R-L}
        & \multirow{-1.25}{*}{GPT-4}
        \\ \hline
        No finetuning & 41.34 & 51.94 \\
        $\mathcal{L}_{\textrm{nll}}$ on $\mathcal{C}$ & 41.81 & 51.94 \\
        $\mathcal{L}_{\textrm{nll}}$ on $\mathcal{C}^\prime$ & 44.52 & \textbf{57.28} \\
        \rowcolor{gray!10} \ours
            & \textbf{44.74} & \textbf{57.28}
            \\
        \thickhline
    \end{tabular}
    \caption{
    Without augmentation (second row) there is a clear performance gap compared to models trained with consistency objectives (third and fourth row).
    }
    \label{tab:mleonly}
\end{table}
}

%% file: tab/tab_scoreaware.tex
\afterpage{
\begin{table}[t]
    \centering
    \fontsize{10.5pt}{12pt}\selectfont
    \begin{tabular}{lcc}
        \thickhline
        & \multicolumn{1}{c}{\multirow{1.25}{*}{MN}}
        & \multicolumn{1}{c}{\multirow{1.25}{*}{MN-IDK}}
        \\
        \cmidrule(r){2-2}
        \cmidrule(r){3-3}
        \multirow{-1.25}{*}{Finetuning Method} 
        & \multicolumn{1}{c}{\multirow{-1.25}{*}{F1}}
        & \multicolumn{1}{c}{\multirow{-1.25}{*}{Acc}}
        \\\hline
        \ours & {58.71} & {98.83}
            \\
        \phantom{0} + Adaptive $\alpha$
            & {59.16} & {98.83}
            \\
        \thickhline
    \end{tabular}
    \caption{
    Effect of dynamically adjusting $\alpha$ based on retriever score.
    }
    \label{tab:scoreaware}
\end{table}
}

%% file: sec/5_conclusion.tex
\section{Conclusion}

We have presented \ours, to balance the tension between
CO (consistency) and RD (rank distillation) objectives in RAG.
For the former,
we augment order-perturbed contexts and add distillation loss for explicit consistency regularization.
For the latter, \ours adaptively chooses desirable degree of perturbation to prevent unlearning valuable prior from the retriever.
\ours consistently outperforms existing methods in
diverse RAG scenarios.

\section*{Limitations}

Whether our findings generalize over diverse models can be further explored.
In addition, the pros and cons of diverse mixing strategies
for an interpolated sample space, 
such as employing another retriever for mix, can be explored; we leave it as future work.

%% file: sec/2_relwork.tex
\section{Related Work}

\subsection{Position bias in long context LLM}

\citet{liu-etal-2024-lost} and similar works have shown that LLMs favor input contexts placed at the beginning or end of the input, a tendency that benchmarks like needle-in-a-haystack\footnote{\href{https://github.com/gkamradt/LLMTest\_NeedleInAHaystack}{github.com/gkamradt/LLMTest\_NeedleInAHaystack}}
aim to assess by testing their ability to locate relevant information (`needle') within long, potentially irrelevant contexts (`haystack').
\citet{an-etal-2024-make} extended this understanding by training models on synthetic data, intentionally
perturbing a position of gold segment and adding random noises.
Similarly, \citet{fu-etal-2024-data-icml} examined continual pretraining of LLMs on long-context data to expand their context window size for retrieving information.

Our distinction is to use position perturbation for a different objective
of data augmentation for consistency training.

\subsection{Data augmentation for consistency}

Pairing a datapoint with a counterfactual applying a small perturbation
has been mainly studied
for robust training on simpler tasks such as classification~\cite{xie-etal-2020-unsupervised-nips}.
To  our knowledge, we are the first to augment a position-perturbed retriever during training and enforce consistency for RAG.

Another related line of work is interpolating two
training instances~\citep{chuang-mroueh-2021-fair-iclr}, which we
extend to define a space
of controlled perturbations for dynamic adaptation
 in Section~\ref{subsec:ours}.

%% file: sec/6_appendices.tex
\section{Implementation Details}

\label{app:impl_details}

\paragraph{MN construction}
For MN data construction, we generally followed the recipe from \citet{an-etal-2024-make}, with the subtle difference that Mixtral was used for question and answer generation.
When preparing the MN dataset following \citet{an-etal-2024-make}, we generally abide by their practices, while using Mixtral as the LLM for question and answer extraction, and employed GPT-4 to verify it.
For the seed corpus, we utilized the same \texttt{realnewslike} subset from the C4 corpus as $\mathcal{C}$. We refer the reader to their original paper for more details.

In addition, to study how LLMs can be trained to refuse to answer when there are insufficient evidence provided, rather than to hallucinate, we split the test set into two settings, answerable and unanswerable:
In the latter, dubbed MN-IDK, the gold segment $s$ that provides the evidence to answer the given question is omitted.
Thus, the model is expected to answer it does not have enough evidence in the contexts to provide the correct answer, or, `I don't know.'

\paragraph{Metrics} The evaluation protocol involving GPT-4 as the judge from \citet{yang-etal-2024-crag} evaluates the correctness of the answer with greater flexibility, compared to the canonical lexical match based metrics, and is known to align better with human judgment.
Also, it penalizes hallucinated response more than simply abstaining.

While other benchmarks considered in this work require shorter answers, expected answers in MN and MN-IDK typically comprise of a few sentences: thus, we report sentence-level F1 score for MN, where GPT-4 was used as a judge in the same manner as the method described above to decide each sentence in the generated answer is supported by the ground-truth (precision), and vice versa (recall).
For MN-IDK, GPT-4 determined whether the model response successfully refused to provide the answer or not, and we reported the accuracy.

\paragraph{Training}
For MS MARCO, HotpotQA and MN, we finetuned Phi-3 3B model on their respective train data: for MS MARCO, we used 20k examples held out from v2.1 dev set for training, and used non-overlapping subset for testing.

For training with \ours on MN, as described in Section~\ref{subsec:ours}, we generated an artificial ranking over the passages by reranking them with a ColBERT variant model from Jina AI,\footnote{\href{https://huggingface.co/jinaai/jina-colbert-v2}{huggingface.co/jinaai/jina-colbert-v2}}\footnote{While our work is completely orthogonal to the choice of retriever, we chose this lightweight model that reportedly perform well across 
several IR benchmarks~\citep{jha-etal-2024-jina}.}
which also provided scores for each passage.
This artificial ranking serves as the opposite extreme of the interpolated perturbation space, $\mathbf{c}^\prime$.

The base model, Phi-3 3B, was trained with LoRA at bf16 precision.
The relevant hyperparameter configuration was as follows:
for LoRA related settings, we used rank of $r=8$, $\alpha=32$, and dropout rate of 0.1.
For general configuration, we used linear decay for scheduling with initial learning rate of 1e-4 and effective batch size of 4; we trained the model for 5 epochs with weight decay of 0.01 applied.
For \ours-specific configuration, we set coefficient for consistency loss strength $\lambda$ as 10 and the noise degree for interpolating contexts $\alpha$ as 0.5 throughout our experiments. 
We leave it as future efforts to search for optimal configuration for these values per different scenarios.

\section{Design of Consistency Loss}

\label{app:first_only}

Using the loss from the first token of the answer only also worked reasonably.
We attribute this to that contribution of the consistency loss terms from earlier time steps, i.e., those from the beginning of the ground-truth, are larger than that of those from later time steps.
The model output probability distribution for time step $t$ defined previously in Section~\ref{subsec:consistency}
is indeed conditioned on the shared prefix of the ground-truth answer $y_{<t}$:
as more tokens in the prefix are conditioned in both sides as $t$ increases, the distribution over the token to be immediately followed $f_t$ would converge, as less and less options would be part of a plausible continuation of the answer.
This results in terms from later $t$ contributing smaller to the total loss $\mathcal{L}_{\textrm{con}}$, which is why dropping all of them but some at the beginning, just one in the extreme case, suffices to regularize the model output.
It is consistent with the findings from previous papers showed that token-level distributional shift between the base and finetuned LLM decreases over time step during decoding~\citep{lin-etal-2024-unlocking-iclr}.

While the benchmarks we have considered generally require rather short responses, it remains to see if this mechanism of using the first time step only for consistency loss computation also work well for long-form answer generation tasks.